\documentclass[11pt,a4paper]{article}

\usepackage[hyperref]{acl2020}
\usepackage{times}
\usepackage{latexsym}

\usepackage{microtype}

\aclfinalcopy 

\usepackage{xcolor}

\newcommand{\lingform}[1]{\textit{#1}}
\newcommand{\err}[1]{{\color{red} \underline{#1}}}
\newcommand{\scare}[1]{'{#1}'}

\newcommand{\lineacross}{\rule{\linewidth}{1pt}}

\title{Shared Task on Evaluating Accuracy}

\author{Ehud Reiter \\
  University of Aberdeen \\
  Aberdeen \\
  UK \\
  \texttt{e.reiter@abdn.ac.uk} \\\And
  Craig Thomson \\
  University of Aberdeen \\
  Aberdeen \\
  UK \\
  \texttt{c.thomson@abdn.ac.uk} \\}

\date{}

\begin{document}
\maketitle
\begin{abstract}
We propose a shared task on methodologies and algorithms for evaluating the accuracy of generated texts, specifically summaries of basketball games produced from basketball box score and other game data.
We welcome submissions based on protocols for human evaluation, automatic metrics, as well as combinations of human evaluations and metrics.
\end{abstract}

\section{Introduction}

Users expect data-to-text NLG systems to generate textual summaries which are accurate.  However, many neural NLG systems in particular generate texts which are factually incorrect.

The most reliable way to assess the accuracy of a generated text is to ask human annotators to carefully fact-check the text.  However this is a time-consuming process.  Our experiences at Aberdeen \citep{INLG20accuracy} show that it can  take an experienced annotator 30 minutes to fact-check a moderately complex 300-word paragraph produced by a neural data-to-text NLG system.  

It would be very useful to the NLG community if we could come up with quicker and easier ways of measuring accuracy which have good correlations with careful fact-checking.   Such methods could be based on less time-consuming human evaluations, such as asking subjects to rate the accuracy of a text on a Likert-type scale \citep{van-der-lee-etal-2019-best}, or on automatic metrics.   However, we should only use such techniques if we feel confident that they have good agreement and correlation with careful high-quality human fact-checking.

The goal of our proposed shared task is to encourage innovative ideas for evaluating accuracy, including both automatic metrics and protocols for human evaluation.  Participants will apply their techniques to summaries of basketball games produced from box score (and other game data) by neural NLG systems.  From the output of three such systems we will compile a corpus of generated texts.

The shared task is unusual because submissions can be protocols for human evaluations as well as computer algorithms (ie, metrics).  The community has limited experience with shared tasks which evaluate protocols, and we hope our experiences will help develop a better understanding of how to undertake such shared tasks, as well as a better understanding of how to evaluate the accuracy of NLG texts.

\section{Organisers}

The organisers are
\begin{itemize}
    \item Ehud Reiter, University of Aberdeen (\texttt{e.reiter@abdn.ac.uk})
    \item Craig Thomson, University of Aberdeen (\texttt{c.thomson@abdn.ac.uk})
\end{itemize}

\section{Task Description}

Participants will be asked to submit one or more submissions which describe either
\begin{itemize}
    \item An evaluation protocol for human subjects which assesses the accuracy of generated texts.  This should include experimental design, guidance on number and type of subjects, and recommended statistical analysis \citep{van-der-lee-etal-2019-best}.  The subjects will have access to data about the game and the teams, and also (if part of the protocol) to a human-authored reference text.
    \item An automatic metric (algorithm) which computes the accuracy of a generated text.  The algorithm will have access to data about the game and the teams, and to a reference text.
    \item A technique which combines human evaluation and automatic metrics.
\end{itemize}
It is acceptable for submissions to give human subjects or metrics access to additional data beyond the box score and other game data used to generate the texts at run-time.  The goal of the shared task is to find statements which are not true in the real world (ie, classic fact-checking), not just statements which disagree with (or are not derivable from) the system run-time data (see Section 3.1 of \citet{INLG20accuracy}).

The output of the evaluation protocol or metric will be a list of mistakes in the text.  Each mistake will be characterised by
\begin{itemize}
    \item Its position in the text (start token and end token).
    \item A category.  We use the following categories, which are based on \citet{INLG20accuracy}
    \begin{itemize}
        \item {\em Incorrect number:}  It does not matter whether the number is spelled out or is in digits.
        \item {\em Incorrect named entity:} This includes people, places, teams, and days of the week.
        \item {\em Incorrect word:} A word which is not one of the above and is incorrect.
        \item {\em Context error:} A phrase which causes an incorrect inference because of context or discourse.
        \item{\em Not checkable:} A statement which can not be checked, either because the information is not available or because it is too time-consuming to check.
        \item {\em Other:}  Any other type of mistake.
    \end{itemize}
\end{itemize}
An example is shown in Figure~\ref{example}.  Note that this example combines fragments from texts produced by several different systems, along with some manual adjustments, in order to illustrate different types of mistakes in a simple way.   Box score and other data for this game is available at	\url{https://www.basketball-reference.com/boxscores/201411050PHO.html} .

\begin{figure}[!b]
\lineacross{}
The Memphis Grizzlies (5-\err{2}) defeated the Phoenix Suns (3 - 2) \err{Monday} 102-91 at the \err{Talking Stick Resort Arena} in Phoenix. The Grizzlies had a \err{strong} first half where they \err{out-scored} the Suns \err{59}-\err{42}. Marc Gasol scored 18 points, \err{leading} the Grizzlies.  \err{Isaiah Thomas added} 15 points, he is \err{averaging 19 points on the season so far}.

\vspace{5mm}
List of errors:
\begin{itemize}
    \item \err{2}: incorrect number, should be 0.
    \item \err{Monday}: incorrect named entity, should be Wednesday.
    \item \err{Talking Stick Resort Arena}: incorrect named entity, should be US Airways Center.
    \item \err{strong}: incorrect word, the Grizzlies did not do well in the first half.
    \item \err{out-scored}: incorrect word, the Suns had a higher score in first half.
    \item \err{59}: incorrect number, should be 46.
    \item \err{42}: incorrect number, should be 52 .
    \item \err{leading}: incorrect word.  Marc Gasol did not lead the Grizzlies, Mike Conley did with 24 points.
    \item \err{Isaiah Thomas added}: context error.  Thomas played for the Suns, but context here implies he played for the Grizzlies and added to their score.
    \item \err{averaging 10 points on the season so far}: not checkable.  This is very hard to check, since data sources report performance per season and per game, not performance up to a particular point in a season.
\end{itemize}
\caption{Example text with error annotations.  Corrections and explanations are not required, but are included here for clarity. Box score data for this game is available at	\url{https://www.basketball-reference.com/boxscores/201411050PHO.html} .}
\label{example}
\lineacross{}
\end{figure}

We will also ask participants to submit estimates of the time required to find mistakes in a text (human time for human evaluations, and CPU/GPU time for metrics).   This is optional, it is not required.  It would, however, be very useful information for the NLG community,

We  also plan to have an \scare{open} track where people can submit ideas for evaluating accuracy on our data set which do not fit into the above framework.

\section{Data}

We will use texts produced by three systems that use basketball box score data: \citet{wiseman-etal-2017-challenges}, \citet{Puduppully1}, and \citet{10.1007/978-3-030-45439-5_5}.   We will carefully fact-check, using the protocol of \citet{INLG20accuracy}, 60 texts (twenty from each system).

The three systems we have chosen all 
explored different ways of modifying the neural architecture.  The system of \citet{wiseman-etal-2017-challenges} defined the Rotowire task and provided initial benchmarks for machine translation systems using copy attention, it is included for this reason.  \citet{Puduppully1} jointly conditioned on a document plan, whilst \citet{10.1007/978-3-030-45439-5_5} used a hierarchical encoder to group attributes (such as statistics) by their respective entities (players/teams).

Other systems in this domain which could be used for evaluation include \citet{puduppully2}, \citet{wang-2019-revisiting}, \citet{gong-etal-2019-table}, and \citet{iso-etal-2019-learning}.  Our aim, however, is to assess how well results produced by the participant's evaluation techniques correlate with the gold-standard fact-checking.  Hence we are looking for a set of systems which generate texts that contain a significant number of accuracy errors, not complete coverage of all systems that generate texts from basketball box score data.    In \citet{INLG20accuracy}, we looked at a number of texts generated by the three systems we have selected.
No text  was error free (the lowest was 7 errors) and the average number of errors per text is about 20.

We will also ask each participant in the shared task to manually fact-check an additional ten texts.  This is optional, but we believe it is very useful for building a better understanding of the task, as well as increasing the amount of training data available.  Note that the protocol of \citet{INLG20accuracy} asks for each text to be fact-checked by three annotators. Since this is expensive, and \citet{INLG20accuracy} report a high-level of inter-annotater agreement, we will accept contributions from participants which have been fact-checked by just one person.  In total, we hope to have 100 fact-checked texts in the training set.

Participants will also have access to all of the texts produced by each of the three systems, along with source box score data and a human-written reference text.

We will create a separate test set of 21 texts which will be manually fact-checked.  The test set will include 7 texts from each of the above three systems.  Each text will be fact checked by 3 annotators following the same process as for the training data.

\section{Evaluation Plans}
We will release the test set (but not the manual fact-checking annotations), and give participants two weeks to apply their techniques to the test set and return the results.  Each mistake will be reported as a position and category, as described above.
We will create a Reported Mistake List (RML) for each annotated text submitted by a participant.

We will then try to align each RML entry with an entry in the gold standard mistake list (GSML).  Match criteria will be applied in order, to each reported mistake (RM) in the RML.  We will mark each match, along with which criteria it was identified under.  In the event of an RM not being found in the GSML
(i.e. a false positive), we will mark `Not found'.  The match criteria are as follows:
\begin{itemize}
    \item Exact match:  First look for a GSML entry which is an exact match to the RML entry.
    \item Same category:  If not found, look for a GSML entry with same category and maximal (non-zero) overlap in position
    \item Different category:  If not found, look for a GSML entry with a different category, with maximal (non-zero) overlap in position
    \item Not found:  If not found, RML entry cannot be aligned with any GSML entry
\end{itemize}

\begin{table}[!h]
    \begin{tabular}{c|c|c|c}
        GSM ID & Token ID`s & Text & Category  \\
        \hline
        GSM-1 & 5-6 & Miami Heat & Name \\
        GSM-2 & 8 & Thursday & Name \\
        GSM-3 & 14-16 & game - high & Word \\
    \end{tabular}
    \caption{GSML for \lingform{`The Denver Nuggets defeated the \err{Miami Heat} on \err{Thursday} . Jamal Murray had a \err{game - high} 30 points .'}.  These are annotated errors, as agreed for our gold standard.  We will use the same tokenization scheme as \citet{wiseman-etal-2017-challenges}.}
    \label{tab:gsml}
\end{table}

\begin{table*}[!h]
    \begin{tabular}{c|c|c|c|l}
        RM ID & Token ID`s & Text & Category & Match to GSML \\
        \hline
        RM-1 & 4-6 & the Miami Heat & Name & GSM-1 Same category (only the determiner differs) \\
        RM-2 & 8 & Thursday & Name & GSM-2 Exact match  \\
        RM-3 & 13-16 & a game - high & Number & GSM-3 Different category (also determiner difference) \\
        RM-4 & 10-11 & Jamal Murray & Name & Not found \\
    \end{tabular}
    \caption{RML for the same text as \autoref{tab:gsml}, but annotated by a submitted method as:  \lingform{`The Denver Nuggets defeated \err{the Miami Heat} on \err{Thursday} . \err{Jamal Murray} had \err{a game - high 30} points .'}.  For each entry in the RML we show here the matching ID (if any) found in the GSML, as well as the match criteria.}
    \label{tab:rml}
\end{table*}

Once we have matched the RSM to the GSML, we will compute a set of scores as follows:
\begin{itemize}
    \item Recall and precision for each category.  In other words, for each category, what percentage of mistakes of this type in GSML were aligned with an RML entry of this category, and vice-versa.
    \item Overall recall and precision (ignoring category).  Looking at RML as a whole, what percentage of entries were successfully aligned with a GSML entry (of any category), and vice-versa.  Given that there are currently many errors (approx 20) in each summary, having a reliable way to detect errors (even without category) would be very useful.
\end{itemize}

\section{Schedule}
We plan on the following schedule.  Note that dates are tentative and may be modified when we know the dates of INLG 2021.
\begin{itemize}
    \item 15 December 2020 (ie, at INLG 2020): announce task, ask for participants
    \item 15 February 2021: deadline for participants to register and (if possible) provide fact-checked stories for training data.
    \item 15 June 2021: submission of techniques (algorithms and protocols).   Test set issued, participants give results on test set within 2 weeks.
    \item 1 August 2021: Results of evaluation computed
    \item INLG 2021: Results presented at INLG, along with posters describing the techniques
\end{itemize}

\bibliography{accuracyeval}
\bibliographystyle{acl_natbib}

\end{document}